\documentclass{article}

\usepackage{PRIMEarxiv}

\usepackage{amsmath}
\usepackage[utf8]{inputenc} 
\usepackage[T1]{fontenc}    
\usepackage{hyperref}       
\usepackage{url}            
\usepackage{booktabs}       
\usepackage{amsfonts}       
\usepackage{nicefrac}       
\usepackage{microtype}      
\usepackage{lipsum}
\usepackage{fancyhdr}       
\usepackage{graphicx}       
\usepackage{multicol}
\usepackage{xcolor}
\usepackage{ulem}
\usepackage{cancel}
\usepackage[linesnumbered,ruled,vlined]{algorithm2e}
\usepackage[numbers]{natbib}
\usepackage{textcomp}
\usepackage{colortbl}
\usepackage{pdfpages}
\usepackage{multirow}
\usepackage{float}
\usepackage{tabularx}
\usepackage{caption}
\usepackage{subfigure}

\title{DGGAN: Degradation Guided Generative Adversarial Network for Real-time Endoscopic Video Enhancement}

\author{
  Handing Xu, Zhenguo Nie \thanks{\textit{Corresponding author}}, Tairan Peng, Xin-Jun Liu \\
  Department of Mechanical Engineering, \\
  State Key Laboratory of Tribology in Advanced Equipment, \\
  Beijing Key Laboratory of Transformative High-end Manufacturing Equipment and Technology \\
  Tsinghua University \\
  Beijing, China\\
  \texttt{\{Handing Xu, Zhenguo Nie\}xhd21@mails.tsinghua.edu.cn, zhenguonie@tsinghua.edu.cn} \\
   \And
  Huimin Pan \\
  Department of Mechanical Engineering \\
  Tsinghua University \\
  Beijing, China\\
}

\begin{document}
\maketitle

\begin{abstract}
Endoscopic surgery relies on intraoperative video, making image quality a decisive factor for surgical safety and efficacy. 
Yet, endoscopic videos are often degraded by uneven illumination, tissue scattering, occlusions, and motion blur, which obscure critical anatomical details and complicate surgical manipulation. 
Although deep learning-based methods have shown promise in image enhancement, most existing approaches remain too computationally demanding for real-time surgical use. 
To address this challenge, we propose a degradation-aware framework for endoscopic video enhancement, which enables real-time, high-quality enhancement by propagating degradation representations across frames. 
In our framework, degradation representations are first extracted from images using contrastive learning. 
We then introduce a fusion mechanism that modulates image features with these representations to guide a single-frame enhancement model, which is trained with a cycle-consistency constraint between degraded and restored images to improve robustness and generalization. 
Experiments demonstrate that our framework achieves a superior balance between performance and efficiency compared with several state-of-the-art methods. 
These results highlight the effectiveness of degradation-aware modeling for real-time endoscopic video enhancement. 
Nevertheless, our method suggests that implicitly learning and propagating degradation representation offer a practical pathway for clinical application. 
\end{abstract}


\section{Introduction}

Minimally invasive surgery (MIS) has become a cornerstone of modern clinical practice, offering reduced surgical trauma, shorter recovery times, and improved postoperative outcomes compared with traditional open procedures. 
Among various MIS techniques, endoscopic surgery plays a particularly critical role, as it enables surgeons to access deep or delicate anatomical regions through narrow working channels with minimal disruption of surrounding tissues. 
For example, in spine surgery, endoscopic approaches have been increasingly adopted for the treatment of conditions such as lumbar disc herniation, spinal stenosis, and degenerative diseases\cite{pan2020percutaneous}. 
Unlike open spine procedures, spine endoscopy relies entirely on intraoperative video as the sole source of visual feedback, making video quality a decisive factor for surgical safety and efficacy. 

However, endoscopic videos are often far from ideal. 
The imaging environment inside the human body presents a variety of challenges: illumination is highly non-uniform due to directional light sources; optical scattering by tissues and fluids degrades image contrast; blood, smoke, and surgical instruments frequently occlude the field of view; and camera motion or limited depth of field can introduce blur. 
These degradations collectively compromise the visibility of fine anatomical structures, hinder accurate surgical manipulation, and may increase the risk of complications. 
In spine endoscopy, where the operative corridor is typically only a few millimeters in diameter and critical neural structures lie in close proximity, such limitations become particularly critical. 
Even subtle degradations in video clarity can obscure vital anatomical cues and complicate intraoperative decision-making.

To address these challenges, video enhancement techniques have been widely investigated. 
Conventional approaches~\cite{swee2020contrast, tan2022endoscope, halidou2023review, yuan2024adaptive} can provide modest improvements in brightness or contrast, but they often fail under severe degradations and may amplify noise or introduce artifacts~\cite{roy2024mathematical}. 
More recently, deep learning-based methods have demonstrated remarkable success in image restoration tasks, including denoising, deblurring~\cite{yang2023capsule}, and super-resolution~\cite{song2022deformable, huang2024deep}, by leveraging large-scale data and learning complex degradation models. 
These techniques have been gradually extended to the domain of endoscopic imaging, with promising results in improving surgical visibility. 

Nevertheless, several critical limitations remain. 
Most existing methods are designed and evaluated in offline settings, where computational efficiency is not a primary concern. 
As a result, they often involve complex models or iterative optimization procedures that are computationally expensive and unsuitable for deployment in the operating room. 
In real surgical environments, enhancement must be performed at video frame rates to provide surgeons with immediate visual feedback. 
The inability of current methods to achieve real-time restoration therefore represents a substantial barrier to clinical adoption.

In this work, we propose a real-time enhancement framework for endoscopic videos. 
We begin by designing a degradation-aware module that estimates an implicit representation of the degradation present in the current frame. 
Building on this representation, we develop a generative single frame enhancement model that trains in a cyclical-consistent manner by degrading the restored image. 
Furthermore, based on the hypothesis that image degradations typically evolve continuously during most surgical procedures, we introduce a Transformer-based degradation propagation module that predicts the degradation representation of subsequent frames from previous ones. 
This design enables the model to directly estimate degradation characteristics only at regular intervals, substantially reducing inference time and making high-quality real-time enhancement of endoscopic videos feasible.
Qualitative and quantitative comparisons with other video or image enhancement methods demonstrate the superior performance and efficiency of the proposed method in enhancing endoscopic videos. 
Our contributions can be summarized as follows:

\begin{itemize}
    \item Inter-frame degradation representation propagation: We proposed a mechanism that propagates degradation representations across frames, enabling real-time video enhancement while maintaining high-quality outputs. 
    \item Single frame enhancement model: We design a single-frame image enhancement model that leverages cyclical consistency to achieve high-quality training. 
    \item Degradation representation fusion mechanism: We introduce a fusion strategy that modulates image features with degradation representations, facilitating degradation-guided feature generation and enhancement. 
\end{itemize}

\section{Related work}

\subsection{Endoscopic video enhancement}
The enhancement of endoscopic video streams is a critical research area focused on correcting dynamic degradations to provide a clear and stable view for clinicians. 
Algorithmic solutions primarily target motion blur and various visual artifacts that compromise video quality. 
Early methods that processed each frame independently were often insufficient, as they failed to account for the temporal nature of video, leading to inconsistencies and suboptimal results. 
Consequently, modern research heavily utilizes the temporal context available in video sequences. 
For instance, in addressing motion blur caused by physiological movement or instrument instability, synthesis-based methods have proven effective. 
These techniques restore sharpness in blurred frames by identifying and transferring high-frequency details from adjacent clear frames, a process demonstrated by~\cite{peng2017endoscopic} using a non-parametric motion model to align the frames accurately.

The advent of deep learning has also marked a paradigm shift in handling image artifacts. 
Of the various artifacts that degrade endoscopic images, specular reflections are particularly problematic, as they can saturate parts of the image and obscure critical underlying tissue. 
While traditional methods relied on filtering and inpainting, deep learning models can now intelligently detect these highlights and reconstruct the missing information. 
Beyond reflections, these models have excelled at removing other common obstructions. The field has evolved from physics-based models, such as the chromaticity-based approach for smoke removal proposed by~\cite{tchaka2017chromaticity}, to more advanced deep learning solutions like improved U-Net architectures that perform real-time desmoking~\cite{lin2021desmoking}. 
Furthermore, to combat the inherent noise in endoscopic sensors, CNN-based blind denoising methods have been developed to handle noise without requiring prior information about its statistical properties~\cite{zou2019cnn, wu2023research}. 
Other specialized models effectively eliminate bubble artifacts in wireless capsule endoscopy videos, which frequently obstruct the view of the gastrointestinal tract~\cite{wang2019reduction}.

\subsection{Real-time image enhancement}
Real-time image enhancement provides immediate visual improvement during live procedures, a crucial capability for on-the-fly diagnosis and intervention. 
This domain is now dominated by deep learning models that offer both the rapid processing speed and the sophisticated analytical power required for clinical application. 
Super-resolution (SR) is a vital technique for improving detail from low-resolution sources, such as capsule endoscopes, or for reducing bandwidth requirements in video streaming. 
The field has progressed rapidly from foundational CNNs to advanced architectures. 
Modern models like EndoL2H use Generative Adversarial Networks (GANs) and spatial attention mechanisms to generate high-fidelity, photorealistic images~\cite{almalioglu2020endol2h}. 
The latest trend involves hybrid models that combine CNNs with transformers to capture both local textural features and long-range contextual dependencies, as seen in HA-VSR~\cite{zhang2023transformer} and E-SEVSR~\cite{hayat2024sevsr}. 
Additionally, saliency-aware and attention mechanisms are being integrated to focus computational resources on diagnostically relevant regions, further boosting performance and efficiency~\cite{hayat2023combined, hayat2024saliency}.

For low-light conditions that can obscure critical anatomical features, Retinex-based deep networks like EIEN have been developed to correct illumination while preserving natural colors and details~\cite{an2022eien}. 
Unsupervised models such as EnlightenGAN offer greater flexibility by not requiring paired training data, which is often difficult to acquire in medical settings~\cite{jiang2021enlightengan}. 
Beyond improving raw image quality, deep learning powers real-time computer-aided detection (CADe) and diagnosis (CADx) systems. 
Functioning as a "second observer," these systems highlight suspicious areas for clinicians. CNN models can detect and classify pathologies like polyps and early-stage cancers in video streams at speeds exceeding 25 frames per second, well within the requirements for real-time clinical use~\cite{guo2020real, yamada2019development}. 
Recent advancements include multi-task models like DSI-Net for joint classification and segmentation~\cite{zhu2021dsi} and robust ensemble learning techniques that improve accuracy and reliability by combining predictions from multiple models~\cite{gunasekaran2023git, nie2024review}.

\section{Methodology}

\subsection{Video enhancement framework}

\begin{figure*}[!ht]
    \centering
    \includegraphics[width=0.95\linewidth]{overall_workflow.png}
    \caption{Framework of the real-time video enhancement. Each frame of the input video sequence undergoes degradation representation modeling and enhancement generation. Specifically, for frames whose index is a multiple of $T_\Delta$, the system invokes the degradation-aware module (DAM) to perform a full estimation, yielding high-precision degradation features. For frames that are not multiples of $T_\Delta$, the degradation representation is rapidly propagated through the degradation representation propagation module (DRPM) along the temporal dimension and subsequently fed into the single frame enhancement model to enhance the current frame. All the modules are detailed in the following sections. }
    \label{fig:overall_workflow}
\end{figure*}

The proposed endoscopic video enhancement framework is illustrated in Fig.~\ref{fig:overall_workflow}. In this framework, each frame of the input video sequence undergoes degradation representation modeling and enhancement generation. 
Specifically, for frames whose index is a multiple of $\Delta_T$, the system invokes the degradation-aware module to perform a full estimation, yielding high-precision degradation features. 
For frames that are not multiples of $\Delta_T$, the degradation representation is rapidly estimated through the DRPM along the temporal dimension and subsequently fed into the generator to enhance the current frame.

This “key-frame—propagated-frame” design enables the framework to maintain accurate degradation representations without performing high-complexity modeling on every frame, thereby significantly reducing the overall computational load. 
More importantly, the propagation mechanism ensures strong temporal consistency and continuity across the video sequence, allowing the system to produce stable enhancement results while satisfying real-time requirements. 
Such a design lays the foundation for online enhancement applications in clinical surgical scenarios.

\subsection{Single frame enhancement model}

We introduce a single-frame image enhancement model that is guided by degradation representation. 
Building upon the CycleGAN framework~\cite{zhu2017unpaired}, our approach leverages cyclical consistency between degraded and restored images to construct the training paradigm. 
Within the generator, we design a dedicated degradation-aware module (DAM) to extract the implicit degradation characteristics of the input image, which are then used to guide the enhancement of the current frame.
This design ensures that the enhancement process is explicitly aware of image degradation, leading to more reliable restoration in complex surgical scenarios. 

\subsubsection{Degradation-aware module}

\begin{figure}[!ht]
    \centering
    \subfigbottomskip=2pt
    \subfigcapskip=-5pt
    \subfigure[]{
        \includegraphics[width=0.6\linewidth]{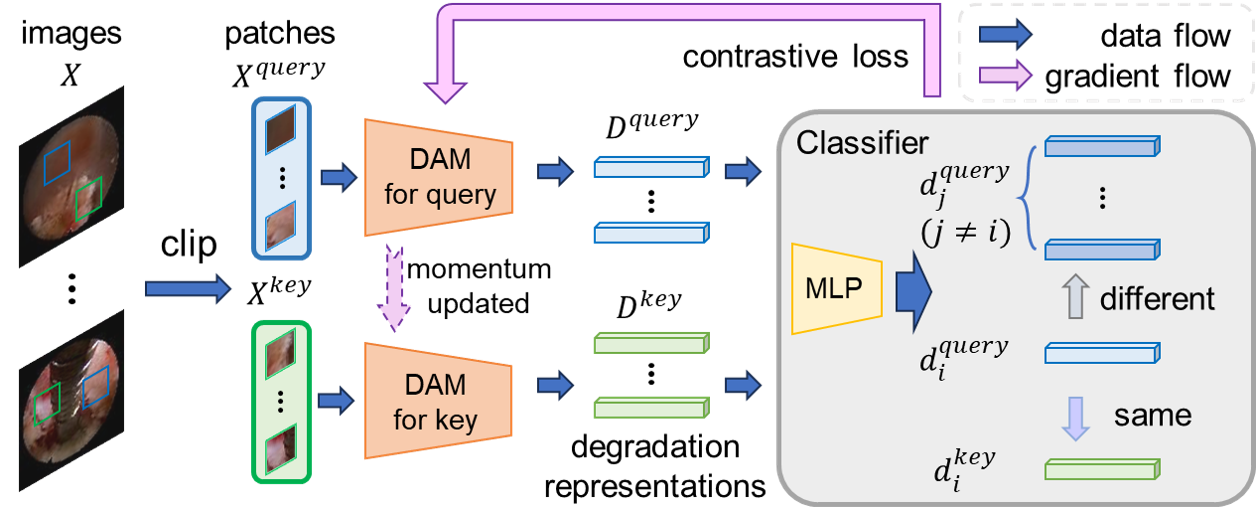}
        \label{fig:degradation_1}}
    \\
    \subfigure[]{
        \includegraphics[width=0.6\linewidth]{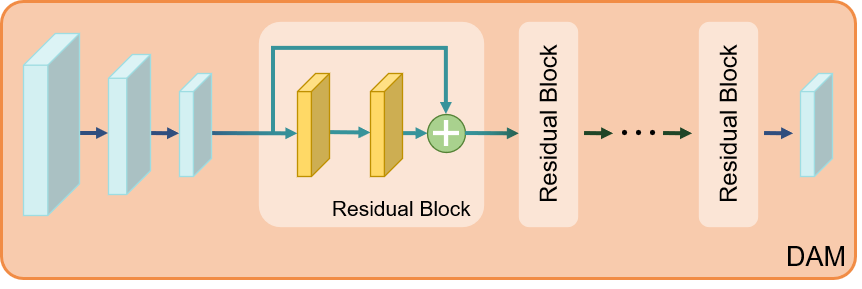}
        \label{fig:degradation_2}}
    \caption{The diagram of the DAM. (a) The DAM training workflow. (b) The structure of the DAM. }
\end{figure}

This module learns to characterize degradation representations through a contrastive learning strategy inspired by MoCo~\cite{he2020momentum, wang2021unsupervised}. 
As illustrated in Fig.~\ref{fig:degradation_1}, given a batch of $N$ images $X$, the assumption is that only representations extracted from the same image should be identical, while those from different images should remain distinct. 
To enforce this property, the degradation-aware module is trained using the InfoNCE loss~\cite{oord2018representation} as \eqref{equ:infoNCE}, which measures the similarity among multiple degradation representations. 

\begin{equation}
    L_{\text{con}} = -\log\frac{exp(d_i^{q}\cdot d_i^{k} / \tau)}{\sum_{j=1}^{N}exp(d_i^{q}\cdot d_j^{k} / \tau)}
    \label{equ:infoNCE}
\end{equation}

To strengthen the expressive capacity of the DAM, we replace conventional stacked convolutional layers with a series of residual blocks, as shown in Fig.~\ref{fig:degradation_2}. 
This architectural refinement allows the encoder to capture richer and more discriminative degradation features. 
Importantly, by producing a more accurate and compact representation of degradation, the DAM reduces the burden on the subsequent image enhancement module. 
As a result, the overall framework not only improves the quality of degradation modeling but also lowers the complexity of the enhancement stage, which is essential for achieving real-time endoscopic video enhancement. 

\subsubsection{Degradation guided enhancement module}

\begin{figure*}[!ht]
    \centering
    \subfigbottomskip=2pt
    \subfigcapskip=-5pt
    \subfigure[]{
        \includegraphics[width=0.95\linewidth]{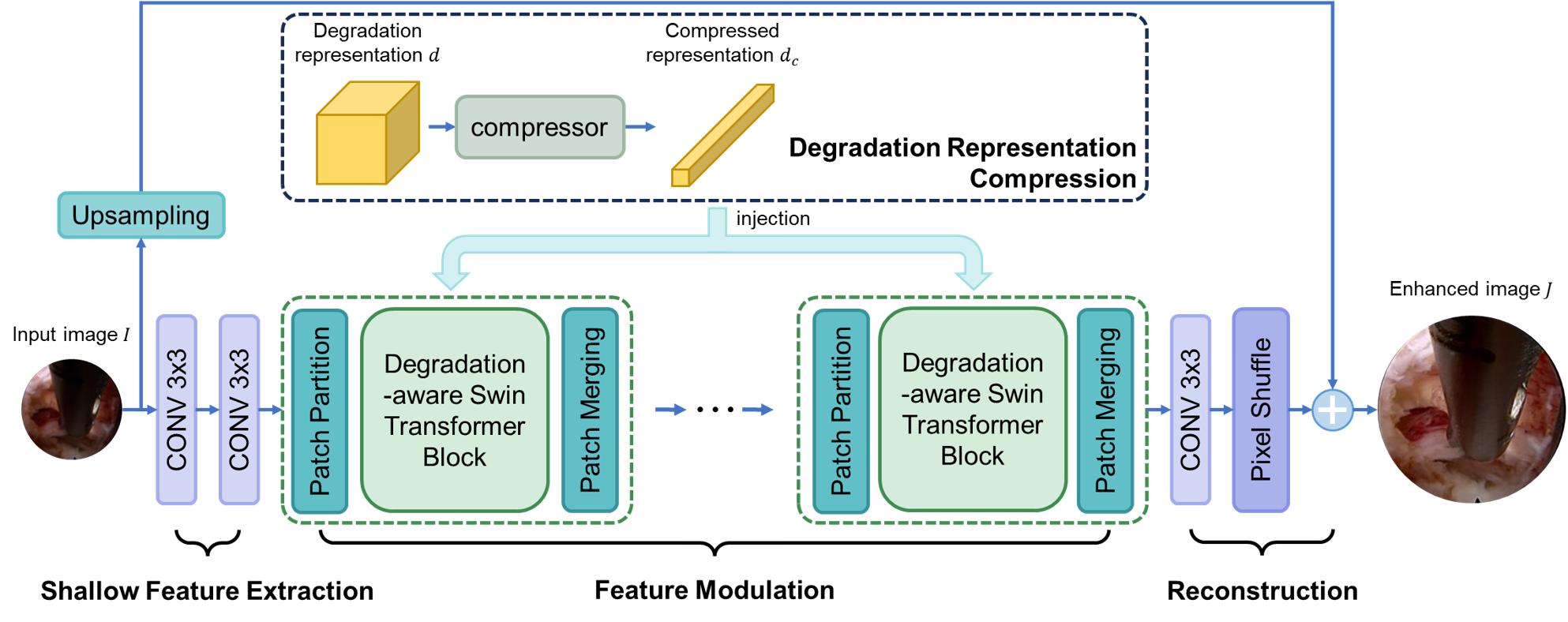}
        \label{fig:generator_1}}
    \\
    \subfigure[]{
        \includegraphics[width=0.59\linewidth]{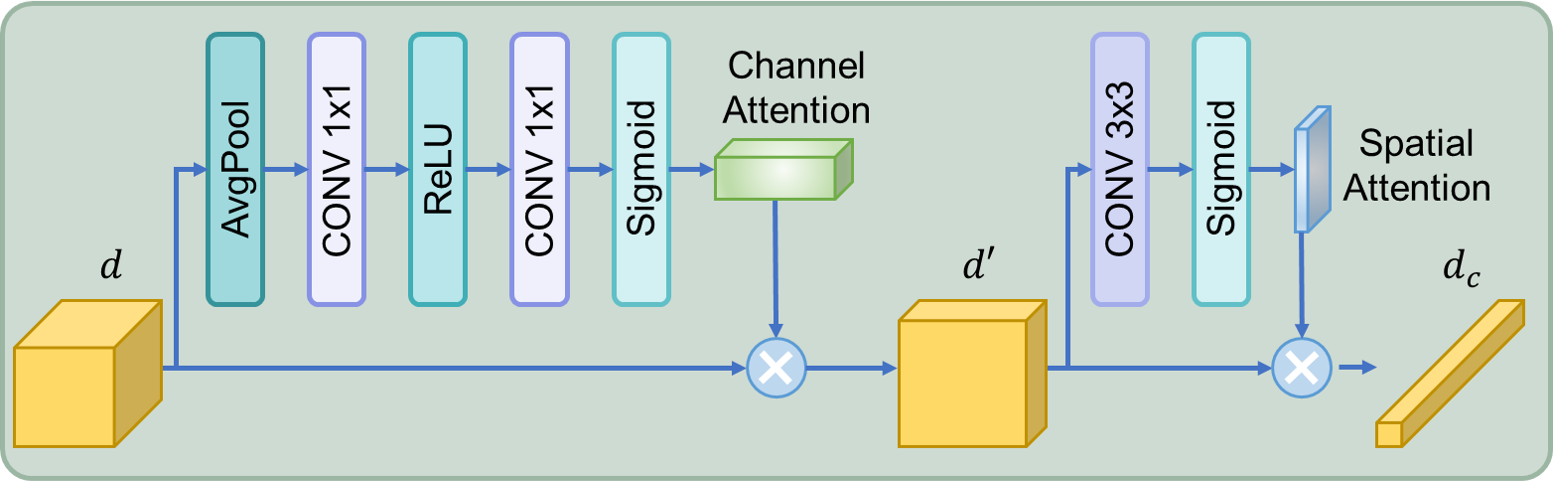}
        \label{fig:generator_2}}
    \subfigure[]{
        \includegraphics[width=0.32\linewidth]{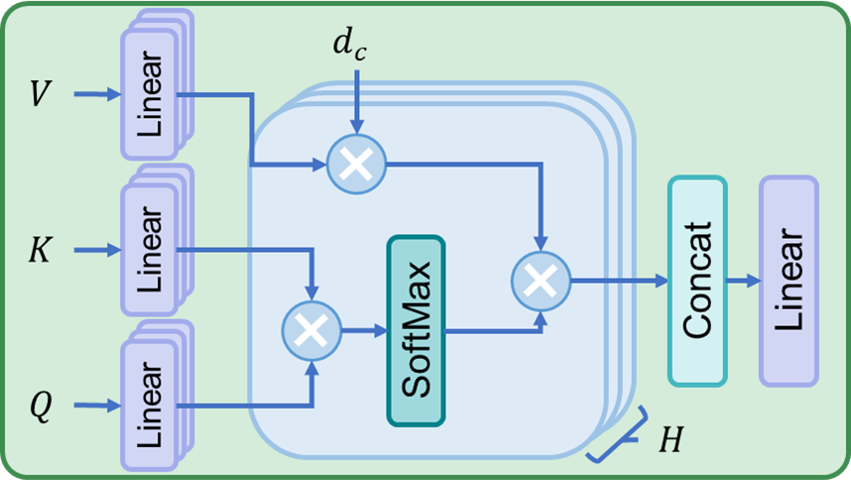}
        \label{fig:generator_3}}
    \caption{Degradation guided enhancement module (DGEM). (a) The basic architecture of the DGEM, which includes four parts: Degradation representation compression, shallow feature extraction, feature modulation, and Reconstruction. (b) The degradation compression block, which implements both channel attention and spatial attention mechanisms. (c) The Degradation-aware Swin Transformer Block, which injects the degradation representation $d_c$ into the value component and modulates the input feature based on multi-head self-attention. }
\end{figure*}

As shown in Fig.~\ref{fig:generator_1}, building on the degradation representation generated by the DAM, we design an degradation guided enhancement module (DGEM) that generates enhanced image in four sequential stages.

\textbf{Degradation representation compression. }
The extracted degradation representation $d$ is high-dimensional and often redundant. 
To obtain a compact yet informative embedding, we introduce both channel attention and spatial attention mechanisms as \eqref{equ:degradation_compression}, which selectively emphasize the most relevant degradation cues while suppressing irrelevant information (Fig.~\ref{fig:generator_2}).

\begin{equation}
    \begin{gathered}
        d' = \sigma(W_{c2}\delta(W_{c1}d)) * d \\
        d_c = \sigma(W_{s1}d') * d'
    \end{gathered}
    \label{equ:degradation_compression}
\end{equation}

\noindent
where $\sigma, \delta$ represent the sigmoid and ReLU activation function separately.

\textbf{Shallow feature extraction. }
We extract low-level structural information from the input image using two stacked $3 \times 3$ convolutional layers. 
This step preserves local edges and textures, which are essential for fine-grained restoration.

\textbf{Feature modulation. }
Inspired by the Swin Transformer~\cite{liu2021swin}, we adopt a shifted-window attention mechanism to capture both local and long-range dependencies. 
Building upon the standard multi-head attention mechanism, we introduce a degradation-aware modulation applied specifically to the value vectors in each attention head, as shown in Fig.~\ref{fig:generator_3}. 
As shown in \eqref{equ:modulation}, by incorporating the degradation representation into the values, the attention module can adaptively emphasize or suppress features according to the estimated degradation in the input image. 

\begin{equation}
    \begin{gathered}
        \hat{V} = V \odot d \\
        \text{MSA}(I) = [\text{Concat}^H_{h=1}(\text{Softmax}(\frac{Q_h K_h^T}{\sqrt{d_h}})\hat{V})]W_o
    \end{gathered}
    \label{equ:modulation}
\end{equation}

\noindent
where $d$ denotes the compressed degradation representation, $Q_h$, $K_h$, and $V_h$ correspond to the query, key, and value of the h\_th attention head, and $d_h$ represents the dimensionality of each haed. 

Unlike modulating the query or key vectors, which would alter the attention distribution itself, modulating the Values directly adjusts the content carried through the attention mechanism, allowing the network to integrate degradation information without disrupting spatial relationships. 
This strategy enables more effective feature aggregation that is guided by the current degradation conditions, ultimately improving the fidelity and robustness of the enhanced image.

\textbf{Reconstruction. }
Finally, the modulated features are passed into a sub-pixel convolution upsampling module, which ensures that the final output benefits from degradation-aware guidance while maintaining high spatial fidelity.

\subsubsection{Cyclical consistency}

\begin{figure}[!ht]
    \centering
    \subfigbottomskip=2pt
    \subfigcapskip=-5pt
    \subfigure[]{
        \includegraphics[width=0.3\linewidth]{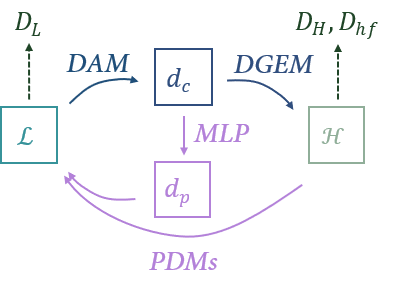}
        \label{fig:cyclical_consistency_1}}
    \subfigure[]{
        \includegraphics[width=0.3\linewidth]{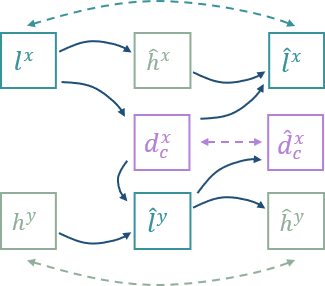}
        \label{fig:cyclical_consistency_2}}
    \caption{Diagram of the cyclical consistency. (a) The dataflow of the whole model. $\mathcal{L} \rightarrow \mathcal{H}$: achieved by the DAM and the DGEM with an additional output $d_c$. $\mathcal{H} \rightarrow \mathcal{L}$: achieved based on PDMs and the regression of $d_p$. (b) Calculation of the cyclical consistency loss. Solid lines indicate the dataflow when generating reconstructed synthetic images, and dashed lines indicate the cyclical consistency loss calculation at both ends. }
\end{figure}

In the CycleGAN framework, style transfer between two image domains is achieved by employing two generators and two discriminators under the constraint of cyclical consistency. 
Inspired by this idea, we construct a cyclical consistent adversarial network between the low-quality and high-quality image domains, denoted as $\mathcal{L}, \mathcal{H}$, respectively. 
As illustrated in Fig.~\ref{fig:cyclical_consistency_1}, the enhancement process $\mathcal{L} \rightarrow \mathcal{H}$ is carried out by the DAM and the DGEM, while the degradation process $\mathcal{H} \rightarrow \mathcal{L}$ is directly implemented based on predefined degradation models (PDMs). 

For the degradation process $\mathcal{H} \rightarrow \mathcal{L}$, as shown in the purple part in Fig.~\ref{fig:cyclical_consistency_1}, multiple lightweight degradation parameter regression modules based on different predefined image degradation models. 
This module estimates explicit degradation parameters $d_p$ from the learned implicit degradation representation, which are then used to synthesize degraded images back. 
Taking the classical image blurring model $\hat{\mathcal{I}} = \sigma \otimes \mathcal{I} + n$ as an example, the regression module can estimate a spatially varying blur kernel $\hat{\sigma}_{i, j}$ together with a noise intensity field $\hat{n}_{i, j}$. 
These estimated parameters are further used to degrade the enhanced images back to the low-quality domain $\mathcal{L}$, thereby enabling training with paired high- and low-quality images. 
Importantly, by grounding the degradation process in physically meaningful models rather than in an unconstrained learned generator, our framework avoids introducing unrealistic artifacts and ensures that the cyclical consistency better reflects the actual imaging degradation. 

In the style of CycleGAN, generator $G$ and the degradation model $D$ aim to produce synthetic high-quality and low-quality images, respectively, whereas the appearance discriminators $D_L, D_H$ compete with the corresponding generators to correctly differentiate the synthetic images from the real images, denoted by \eqref{equ:adversarial_loss_1} and \eqref{equ:adversarial_loss_2}. 
For simplicity, we use $G_H$ to denote the combination of DAM and DGEM, and $G_L$ to denote PDMs. 
To encourage the generation of high-quality enhanced images, we extend the standard two-discriminator framework by adding a high-frequency discriminator $D_{hf}$ that focuses specifically on the high-frequency of the high-quality image domain. 
This design aims to guide the generator in preserving fine-grained details and texture information, and its corresponding adversarial loss is defined as \eqref{equ:adversarial_loss_3}.

\begin{align}
    &\begin{aligned}
        L_{\text{adv}}(G_H, D_H) = &\mathbb{E}_{x_h \sim p(x_h)}[\log(D_H(x_h))] + \\
        &\mathbb{E}_{x_l \sim p(x_l)}[\log(1-D_H(G_H(x_l)))]
    \end{aligned} \label{equ:adversarial_loss_1} \\
    &\begin{aligned}
        L_{\text{adv}}(G_L, D_L) = &\mathbb{E}_{x_l \sim p(x_l)}[\log(D_L(x_l))] + \\
        &\mathbb{E}_{x_h \sim p(x_h)}[\log(1-D_L(G_L(x_h)))]
    \end{aligned} \label{equ:adversarial_loss_2} \\
    &\begin{aligned}
        L_{\text{adv}}(G_H, D_{hf}) = &\mathbb{E}_{x_h \sim p(x_h)}[\log(D_{hf}(F_{hp}(x_h)))] + \\
        &\mathbb{E}_{x_l \sim p(x_l)}[\log(1-D_{hf}(F_{hp}(G_H(x_l))))]
    \end{aligned} \label{equ:adversarial_loss_3}
\end{align}

\noindent
where $F_{hp}$ denotes a highpass filter. 

As shown in Fig.~\ref{fig:cyclical_consistency_2}, the cyclical consistency loss is defined as \eqref{equ:cyclical_loss}, which forces the reconstructed synthetic images $G_L(G_H(L))$ and $G_H(G_L(H))$ to be identical to their respective inputs $L$ and $H$. 

\begin{equation}
    \begin{aligned}
        L_{\text{cyc}}(G_H, D) = &\mathbb{E}_{x_l \sim p(x_l)}[\parallel G_L(G_H(x_l)) - x_l \parallel_1] + \\
        &\mathbb{E}_{x_h \sim p(x_h)}[\parallel G_H(G_L(x_h)) - x_h \parallel_1] + \\
        &\mathbb{E}_{x_l, x_h \sim p(x_l), p(x_h)}[\parallel D(G_L(x_h)) - D(x_l) \parallel_1]
    \end{aligned}
    \label{equ:cyclical_loss}
\end{equation}

\noindent
where the third term denotes the cyclical consistency of the degradation representation, and $D$ in it represents the DAM. It should be mentioned that the degradation representation $d_c$ can be only produced by the DAM within $G_H$. Consequently, in the term $G_L(x_h, d_c^h)$, the degradation representation $d_c^h$ can only be replaced with $d_c^l$ that derived from $G_H(x_l)$, which means that the degradation representation cycle contains only a single term. 

\subsection{Degradation representation propagation module}

Although the aforementioned single-frame image enhancement model can achieve high-quality endoscopic image restoration, its inference efficiency is insufficient to meet the real-time requirements of intraoperative observation. 
To address this, we balance the complexity of different model components against overall performance. 
Specifically, we manually adjust the number of residual blocks and Swin Transformer blocks to adjust the number of parameters and floating-point operations (FLOPs) for each module (detailed in Table \ref{tab:param_FLOPs}), concentrating the primary computational cost in the degradation representation extraction stage.

\begin{table}[!ht]
    \renewcommand\arraystretch{1.3}
    \centering
    \caption{Number of parameters and FLOPs for each part of DGGAN.}
    \begin{tabular}{cccc}
        \hline
        \multicolumn{2}{c}{}                                                                & Params(M) & GFLOPs \\ \hline
        \multicolumn{2}{c}{DAM}                                                             & 4.33      & 29.75  \\
        \multicolumn{2}{c}{DGEM}                                                            & 0.39      & 24.65  \\ \cline{2-2}
        & \begin{tabular}[c]{@{}c@{}}Degradation Representation \\ Compression\end{tabular} & 0.066     & 1.31   \\
        & Shallow Feature Extraction                                                        & 0.0018    & 0.11   \\
        & Feature Modulation                                                                & 0.13      & 8.22   \\
        & Reconstruction                                                                    & 0.19      & 15.01  \\ \hline
    \end{tabular}
    \label{tab:param_FLOPs}
\end{table}

Based on these adjustments, we further introduce a degradation representation propagation module (DRPM) to fully exploit temporal correlations across video frames. 
This module leverages the degradation representations of previous frames as implicit priors, propagating and updating them along the temporal dimension to efficiently estimate the degradation representations of future frames. 
Compared with modeling each frame independently, this propagation mechanism significantly reduces redundant computations by avoiding repeated processing of highly similar information across adjacent frames, thereby lowering overall computational complexity and ensuring real-time enhancement. 
Moreover, by maintaining inter-frame consistency of the degradation representations, the module enhances the stability and temporal coherence of degradation modeling across sequences.

\begin{figure}[!ht]
    \centering
    \includegraphics[width=0.6\linewidth]{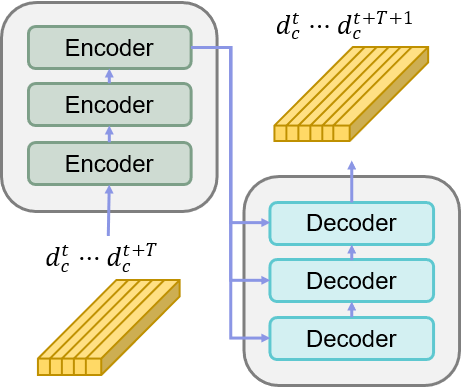}
    \caption{The architecture of the DRPM. }
    \label{fig:DRPM}
\end{figure}

In implementation, the DRPM is built upon a Transformer architecture as shown in Fig.~\ref{fig:DRPM}. Unlike conventional recurrent neural networks (RNNs) or 3D convolutional networks (3D-CNNs), the self-attention mechanism in Transformers enables flexible modeling of dependencies over a long temporal range, overcoming the locality constraints of convolution or the limited temporal span of RNNs. 
Through multi-head attention, the module can simultaneously capture multi-scale temporal correlations in the degradation representations, accurately modeling both local dynamics between adjacent frames and long-range dependencies across distant frames. 
As a result, employing a Transformer not only improves the accuracy and robustness of degradation representation estimation but also preserves temporal consistency and overall stability in the enhanced video sequence.

Importantly, to further improve the efficiency of degradation propagation while maintaining effectiveness, the module does not propagate the full degradation representation $D$ directly. 
Instead, it uses a compressed representation $\hat{D}$ as the temporal modeling unit, which is not directly illustrated in Fig.~\ref{fig:overall_workflow}. 
This strategy reduces feature dimensionality and computational redundancy, while also enhancing the model's generalization and real-time performance, providing an efficient degradation-aware support for continuous-frame endoscopic image enhancement.

\subsection{Training strategy}

\begin{algorithm}[]
\caption{Training strategy}
\label{alg:train_strategy}
\DontPrintSemicolon 
\KwIn{A dataset of high quality endoscopic images $X_h$, A dataset of low quality endoscopic videos $X_l$, A predefined degradation model $\text{PDM}$}
\SetAlgoNoLine
\For{epoch in $0, N_d$} {
    Sampling $x_h$ from $X_h$\;
    Generate $\hat{x}_l$ by degrading $\text{PDM}(x_h)$\;
    Train DAM by minimizing L{\text{con}}\;
}
\For{epoch in $N_d, N_d + N_s$} {
    Sampling $x_h$ from $X_h$\;
    Sampling $x_l$ from $X_l$ as images\;
    Train $G_L, G_H$ by minimizing $L_{\text{adv}} + L_{\text{cyc}}$\;
    Train $D_L, D_H, D_{hf}$ by maximizing $L_{\text{adv}}$\;
}
\For{epoch in $N_d + N_s, N$} {
    Sampling $x_h$ from $X_h$\;
    Sampling $x_l^t, ..., x_l^{t+T}$ from $X_l$ as videos\;
    Replace DAM with DRPM\;
    Train DRPM by minimizing $L_{\text{adv}} + L_{\text{cyc}}$\;
    Train $D_L, D_H, D_{hf}$ by maximizing $L_{\text{adv}}$\;
}
\end{algorithm}

The training strategy is detailed in Algorithm \ref{alg:train_strategy}. 
In first $N_d$ epochs, the model first pre-trains the DAM using the contrastive loss $L_{con}$ on paired endoscopic images generated by artificial degradations, enabling it to capture effective degradation representations. 
After this stage, we switch to real-world data, training the model on unpaired low-quality and high-quality endoscopic images under the CycleGAN paradigm. 
During adversarial learning, to train the generators, the associated objective is minimized while to train the discriminators, the associated objective is maximized. 
Please note that when the generators are optimized, the weights of the discriminators are fixed and vice versa. 
At each step, we conduct following computations including forward propagation, loss calculation, backward propagation and updating weights of associated components.
Finally, for video datasets, non-key frames are processed by replacing the DAM in the single-frame enhancement model with the DRPM. 
The training is then conducted using the same loss functions as described above, which guide the DRPM to propagate and generate reasonable degradation representations. 

The design of this two-stage dataset utilization scheme is motivated by two key considerations.
On one hand, acquiring large-scale paired endoscopic data in clinical practice is extremely challenging, particularly for image pairs exhibiting diverse degradation types, which has a significant impact on the generalization capability of the DAM. 
On the other hand, models trained solely on artificially degraded pairs often generalize poorly to real surgical videos. 
By first initializing the degradation representation module with artificial pairs and then adapting the whole model with real unpaired data, our approach combines the advantages of both worlds and ensures better performance in realistic surgical environments.

\section{Experiments and results}

In this section, we first evaluate the performance of the proposed method on the SCARED dataset~\cite{allan2021stereo} and compare it with several state-of-the-art deep learning-based approaches. 
We then assess the degradation-awareness capability of the proposed module, demonstrating the effectiveness of the learned degradation representations. 
Furthermore, we evaluate the performance of our method under different degradation patterns across various surgical scenarios. 
To gain deeper insights into the contribution of each component, we conduct ablation studies. 
Finally, we construct a simulated surgical environment to examine the real-time performance of the proposed approach.
Code is available at: https://github.com/ilk123/DGGAN.

\subsection{Data collection and implementation details}

\subsubsection{Datasets and preprocessing}

In this study, we construct two datasets to support model training and evaluation. 
For the high-resolution images, we adopt the SCARED dataset. 
Although originally designed for monocular depth estimation, we select $2,295$ clear and high-quality left-camera images to build a high-quality endoscopic image dataset.
For the low-resolution videos, we utilize the spinal endoscopic surgery dataset (referred as the SES dataset) collected at Peking University People's Hospital, from which we extract $240$ clips, each consisting of $120$ frames. 
The choice of this dataset is motivated by the fact that Percutaneous Endoscopic Lumbar Discectomy (PELD) is performed in a narrow operative field filled with irrigation fluid and image is degraded severely, which may post different levels of challenge to our method. 

Specifically, the resolution of the SCARED dataset is $1280 \times 1024$, while the spinal endoscopic dataset has resolutions of $720 \times 1080$ or $1920 \times 1080$. 
In order to reduce the redundant areas of the images, such as pure black areas and inner wall of the working channel, we crop the images into $320 \times 320$ patches to construct the low-quality endoscopic dataset. 
Finally, the SCARED dataset and the SES dataset are split into training, validation, and testing sets following a 7:2:1 ratio.

\subsubsection{Implementation details}

We train our model following the procedure outlined in Algorithm \ref{alg:train_strategy}. 
Throughout the training process, we adopt the Adam optimizer with parameters set to $\beta_1 = 0.9$ and $\beta_2 = 0.999$. 
During the training of the DAM, the learning rate is set to $5e-5$. 
For the training of the single-frame enhancement model, the learning rate of G and D are set to $5e-5$ and $2e-4$, separately. 
For the DRPM, we configure the time interval for DAM-based degradation estimation to $15$ frames, with the same learning rate as the previous stage. 
All experiments are conducted using two NVIDIA RTX 3090Ti GPUs for training, while a single NVIDIA RTX 3090Ti is employed for model testing.

\subsection{Enhancement performance and efficiency of DGGAN}

We compare our method with several open-source image enhancement algorithms, including degradation-specific approaches as well as two general-purpose methods (Restormer~\cite{zamir2022restormer} and SwinIR~\cite{liang2021swinir}). 
For our framework, we evaluate both the two variants: the degradation-guided enhancement model based on DAM (DGGAN-DAM) and the degradation-guided enhancement model based on DRPM (DGGAN-DRPM).

We first conduct experiments on the SCARED dataset, where the low-quality test images are generated by applying degradations based on multiple degradation models, as defined below. 
Since the ground truth is available in this setting, we adopt PSNR~\cite{kim2016accurate} and SSIM~\cite{lai2017deep} as full-reference metrics to evaluate image quality. 
We then conduct experiments on the spinal endoscopic dataset, where no ground truth is available. 
In this case, we employ no-reference metrics, including NIQE~\cite{mittal2012making} and PIQE~\cite{venkatanath2015blind}, to assess the performance of different models.

In addition, to evaluate the computational efficiency of each model, we reported two key complexity measures-the number of parameters and the floating point operations (FLOPs)-following common practice in super-resolution studies.
To measure the computational efficiency of each model, we calculate two essential properties: the number of parameters and FLOPs (Floating-point operations per second). 
The number of parameters in a model describes its size, with fewer parameters indicating a more lightweight model. 
FLOPs can describe the computational complexity of the model, and have become a widely used metric to measure the efficiency of models. 

\subsubsection{Degradation models}

On the SCARED dataset, we apply four degradation types frequently observed in endoscopic surgery: noise, motion blur, low illumination, and electrocautery smoke. 
For each type, the corresponding degradation model is implemented as summarized in Table \ref{tab:degradation_type}, and during the training of our DGGAN model, the associated parameters are regressed to support the degradation process.

\begin{table}[!ht]
    \renewcommand\arraystretch{1.3}
    \centering
    \caption{The corresponding models for different degradation types. }
    \begin{tabular}{c|c}
        \hline
        Degrade Type & Degrade Model     \\ \hline
        Random Noise & $I(x)=J(x)+n(x)$     \\ \hline
        Motion Blur  & $I(x)=k(x) \otimes J(x)$ \\ \hline
        Low Light    & $I(x)=J(x)\cdot L(x) + n(x)$                \\ \hline
        Smoke        & $I(x)=J(x)t(x) + A(1-t(x))$                \\ \hline
        SES          & referred as to \eqref{equ:deg_model} \\ \hline              
    \end{tabular}
    \label{tab:degradation_type}
\end{table}

It should be mentioned that although explicit degradation models and their parameters are calculated, we deliberately choose not to use these parameters as direct supervision signals. 
Instead, our framework learns to encode and exploit degradation representations that implicitly capture the degradation characteristics. 
This design enables the model to move beyond parameter-specific regression and achieve greater robustness and generalization across diverse degradations.

For the spinal endoscopy dataset, we further construct a degradation representation model \eqref{equ:deg_model} tailored to real surgical scenarios, taking into account four major aspects: low resolution, blurring, low illumination, and underwater imaging effects.

\begin{equation}
    I(x) = (\beta (\alpha J(x) \otimes k)^\gamma T(x) + A(1-T(x)))\downarrow_s + n
    \label{equ:deg_model}
\end{equation}

\subsubsection{Results on SCARED dataset}

\begin{figure*}[!ht]
    \centering
    \includegraphics[width=0.9\linewidth]{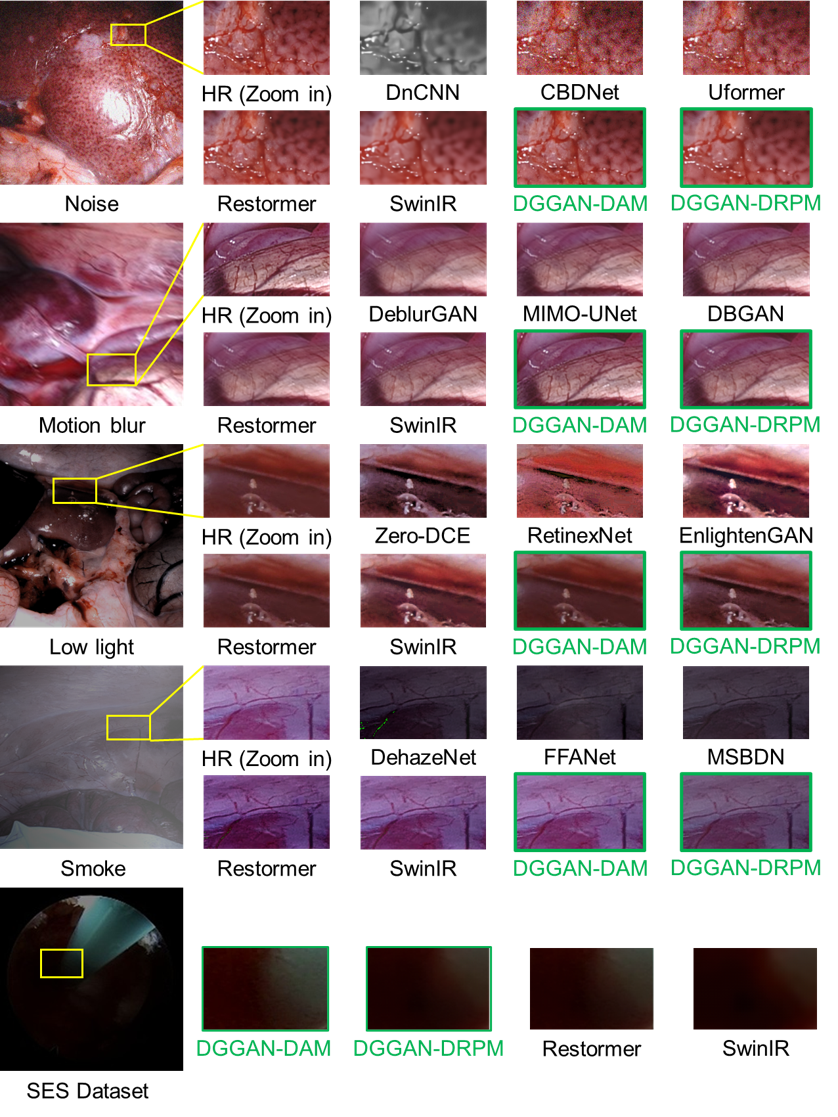}
    \caption{Qualitative comparison of enhancement results on degraded endoscopic datasets. The first four rows correspond to the SCARED dataset under four degradation types: noise, motion blur, smoke, and low light. The last row shows results on our SES dataset.}
    \label{fig:image}
\end{figure*}

The performance of multiple models on SCARED dataset are shown in Table \ref{tab:image_quality} and Fig.~\ref{fig:image}. It is worth noting that since our DGGAN framework incorporates an inherent super-resolution step, for competing methods without a upsampling module, we adopt a strategy of first applying enhancement and then bicubic interpolation for resolution restoration. 
While this inevitably reduces their final performance, it keeps their inference efficiency comparable.

For the denoising task, we compare against DnCNN~\cite{zhang2017beyond}, CBDNet~\cite{guo2019toward}, and Uformer~\cite{wang2022uformer}. 
As denoising represents the simplest enhancement scenario, the performance of all models is remarkable except DnCNN, since it was trained for Grayscale image denoising. 
Notably, our DGGAN-DRPM achieves results only slightly lower than Restormer and SwinIR, while requiring just $30\%$ and $7.3\%$ of their FLOPs, respectively.

For the deblurring task, we select DeblurGAN~\cite{kupyn2018deblurgan}, MIMOUNet~\cite{cho2021rethinking}, and DBGAN~\cite{zhang2020deblurring}. 
In this case, DGGAN-DAM achieve the best overall performance, surpassing Restormer by $0.1083$ in SSIM, while DGGAN-DRPM reaches 0.7785 SSIM, ranking just behind DGGAN-DAM and SwinIR.

For the low-light enhancement task, we compared against Zero-DCE~\cite{guo2020zero}, RetinexNet~\cite{wei2018deep}, and EnlightenGAN~\cite{jiang2021enlightengan}. 
Here, DGGAN-DAM nearly matches Restormer's performance (a marginal SSIM drop of $0.0038$), while DGGAN-DRPM rankes third overall, outperforming SwinIR by $0.0327$.

For the electrocautery smoke removal task, we selected DehazeNet~\cite{cai2016dehazenet}, FFANet~\cite{qin2020ffa}, and MSBDN~\cite{dong2020multi}. 
Again, DGGAN-DAM achieves the best results, though DGGAN-DRPM underperformes compared to Restormer and SwinIR. 
We attribute this to the inherently stochastic nature of smoke diffusion, which limits the ability of the DRPM to effectively propagate degradation representations across frames. 

Overall, these results highlight that even when compared against task-specific enhancement models under comparable efficiency settings, our framework consistently achieves an outstanding balance between computational efficiency and enhancement quality. 
In particular, for each degradation type, we deliberately choose relatively lightweight baseline models to ensure similar inference efficiency, thereby making the performance gains of our method more directly attributable to its superior balance between efficiency and effectiveness.

\begin{table*}[!ht]
    \renewcommand\arraystretch{1.3}
    \centering
    \caption{Performance comparison in terms of PSNR (dB), SSIM, parameter counts, FLOPs, and running time (s) of the proposed DGGAN with the SOTAs on the SCARED dataset. }
    \begin{tabular*}{\hsize}{@{}@{\extracolsep{\fill}}ccccccc@{}}
        \hline
        Degradation                   & Method       & PSNR(dB)$\uparrow$ & SSIM$\uparrow$  & Parameters(M)$\downarrow$ & GFLOPs$\downarrow$  & Running time(s)$\downarrow$ \\ \hline
        \multirow{7}{*}{Random Noise} & DnCNN        & 15.50     & 0.7265 & 0.56       & 28.11  & 0.0205   \\
                                      & CBDNet       & 28.98     & 0.8185 & 4.43       & 30.84  & 0.0385   \\
                                      & Uformer      & 30.96     & 0.8534 & 20.63      & 41.09  & 0.0419   \\ \cline{2-7} 
                                      & Restormer    & 31.52     & 0.8710 & 26.13      & 108.0  & 0.5271   \\
                                      & SwinIR       & \underline{32.76}     & \underline{0.8902} & 10.89      & 711.9  & 11.242   \\
                                      & DGGAN-DAM    & \textbf{33.21}     & \textbf{0.9057} & 4.72       & 54.41  & 0.1459   \\
                                      & DGGAN-DRPM   & 31.03     & 0.8678 & 0.63       & 32.40  & 0.0291   \\ \hline
        \multirow{7}{*}{Motion Blur}  & DeblurGAN    & 12.87     & 0.4879 & 6.06       & 26.88  & 0.0738   \\
                                      & MIMOUNet     & 17.59     & 0.6783 & 6.81       & 51.26  & 0.0083   \\
                                      & DBGAN        & 18.54     & 0.7121 & 11.59      & 74.35  & 0.1301   \\ \cline{2-7} 
                                      & Restormer    & \underline{23.01}     & \underline{0.8127} & 26.13      & 108.0  & 0.6159   \\
                                      & SwinIR       & 22.97     & 0.8094 & 10.89      & 711.9  & 10.855   \\
                                      & DGGAN-DAM    & \textbf{26.86}     & \textbf{0.9210} & 4.72       & 54.41  & 0.1421   \\
                                      & DGGAN-DRPM   & 20.85     & 0.7785 & 0.63       & 32.40  & 0.0305   \\ \hline
        \multirow{7}{*}{Low Light}    & Zero-DCE     & 19.11     & 0.7085 & 0.024      & 1.178  & 0.0018   \\
                                      & RetinexNet   & 19.16     & 0.7838 & 0.56       & 16.78  & 2.4734   \\
                                      & EnlightenGAN & 19.26     & 0.7755 & 54.41      & 127.45 & 0.1187   \\ \cline{2-7} 
                                      & Restormer    & \textbf{28.83}     & \textbf{0.8657} & 26.13      & 108.0  & 0.5764   \\
                                      & SwinIR       & 25.49     & 0.8018 & 10.89      & 711.9  & 12.843   \\
                                      & DGGAN-DAM    & \underline{28.01}     & \underline{0.8619} & 4.72       & 54.41  & 0.1399   \\
                                      & DGGAN-DRPM   & 26.54     & 0.8345 & 0.63       & 32.40  & 0.0314   \\ \hline
        \multirow{7}{*}{Smoke}        & DehazeNet    & 16.86     & 0.6899 & 0.02       & 0.391  & 0.5237   \\
                                      & FFANet       & 23.05     & 0.7311 & 4.68       & 231.2  & 0.3891   \\
                                      & MSBDN        & 19.01     & 0.6529 & 28.71      & 18.81  & 0.1031   \\ \cline{2-7} 
                                      & Restormer    & \underline{27.37}     & \underline{0.8346} & 26.13      & 108.0  & 0.6038   \\
                                      & SwinIR       & 26.79     & 0.7981 & 10.89      & 711.9  & 11.745   \\
                                      & DGGAN-DAM    & \textbf{28.18}     & \textbf{0.8553} & 4.72       & 54.41  & 0.1407   \\
                                      & DGGAN-DRPM   & 27.13     & 0.7823 & 0.63       & 32.40  & 0.0276   \\ \hline
    \end{tabular*}
    \label{tab:image_quality}
\end{table*}

\subsubsection{Results on SES dataset}

\begin{table}[!ht]
    \renewcommand\arraystretch{1.3}
    \centering
    \caption{The Evaluation results of NIQE/PIQE on our SES dataset. }
    \begin{tabular}{ccccc}
        \hline
                             & Method      & NIQE$\downarrow$ & PIQE$\downarrow$ & Parameters$\downarrow$ \\ \hline
        \multirow{4}{*}{SES} & Restormer   & 4.32  & \underline{15.79} & 26.13M      \\
                             & SwinIR      & 5.03  & 18.21 & 10.89M      \\
                             & DGGAN-DAM   & \textbf{3.62}  & \textbf{13.79} & 4.72M       \\
                             & DGGAN-DRPM  & \underline{3.98}  & 16.82 & 0.63M       \\ \hline
    \end{tabular}
    \label{tab:SES_quality}
\end{table}

We conduct evaluations on our SES dataset subsequently. 
Although the performance can only be assessed using no-reference quality metrics, specifically NIQE and PIQE, the results in Table \ref{tab:SES_quality} and Figure \ref{fig:image} clearly demonstrate that our method significantly outperforms both Restormer and SwinIR. 
This finding highlights the strong capability of DGGAN in handling complex and diverse degradation scenarios. 
More importantly, in real surgical environments, the challenges differ markedly from those in general image enhancement tasks. 
While patient-specific anatomical and physiological conditions may vary, the types of visual degradations that occur during a given minimally invasive surgery tend to remain relatively consistent. 
For example, in spinal endoscopy surgery, we can systematically characterize the degradation process with a degradation model such as the one formulated in \eqref{equ:deg_model}. 
Leveraging this insight, DGGAN is able to achieve reliable and real-time enhancement performance. 
This implies that, for a specific minimally invasive surgical procedure, it is sufficient to identify and model the major degradation types that are likely to occur. 
Once this is done, a dedicated enhancement model can be efficiently trained, enabling high-quality and real-time endoscopic video enhancement that directly benefits intraoperative visualization and surgical safety.

\subsection{Degradation Representation performance}

Degradation representation learning is a key component of our DGGAN framework, as it enables the model to obtain an implicit estimation of endoscopic image degradations, which in turn guides the overall training process. 
To validate the effectiveness of this learned representation, we performed a visualization analysis. 

\begin{figure*}[!ht]
    \centering
    \includegraphics[width=0.95\linewidth]{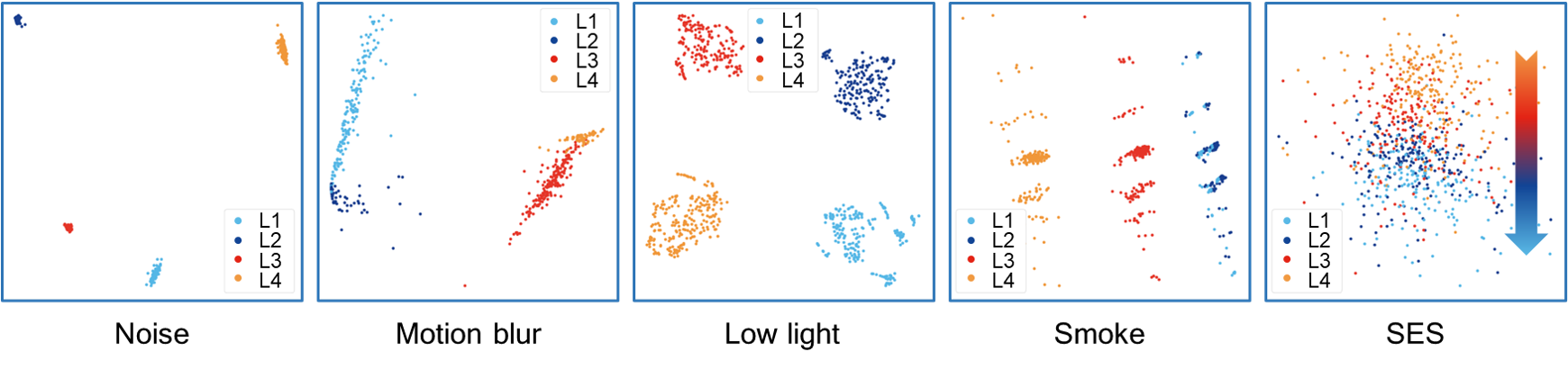}
    \caption{The degradation representation performance of the DAM under different degradation types. }
    \label{fig:degradation_PCA}
\end{figure*}

Specifically, based on the degradation types defined in Table \ref{tab:degradation_type}, we apply four levels of artificial degradations to high-quality endoscopic images, denoted as L1-L4. 
These degraded images are then fed into the DAM to extract their degradation representations. 
To facilitate interpretation, we further project the high-dimensional representations into a two-dimensional space using PCA~\cite{abdi2010principal} for visualization. 
As shown in Fig.~\ref{fig:degradation_PCA}, the DAM successfully learns meaningful degradation features: different degradation types, such as noise and motion blur, form clearly separable clusters. 
Even in the case of more complex degradations typical of spinal endoscopic surgery, as modeled in \eqref{equ:deg_model}, the DAM is still able to capture and distinguish degradation severity to a certain extent, as indicated by the arrow. 
These results confirm that the DAM provides effective and discriminative degradation representation learning, which forms the basis for robust enhancement performance in diverse surgical scenarios.

\subsection{Ablation Study}

\subsubsection{Effectiveness of the degradation representation}

\begin{figure}[!ht]
    \centering
    \includegraphics[width=0.6\linewidth]{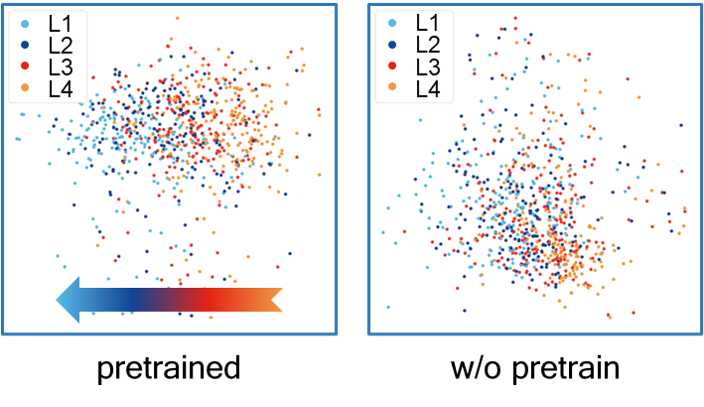}
    \caption{The degradation representation performance on SES dataset with or without pretraining. }
    \label{fig:degradation_representation}
\end{figure}

In this study, we investigate the role of degradation representation in our framework. 
To this end, we remove the pretraining stage of the DAM module in Algorithm \ref{alg:train_strategy} (referred to as DGGAN w/o p), which to some extent simulates the vanish of degradation representation learning, and compare it with our full model. 
The visualization results of the learned degradation representations and the corresponding enhancement performance on the SES dataset are shown in Fig. \ref{fig:degradation_representation} and Table \ref{tab:degradation_representation}, respectively. 
Without pretraining, the model almost completely loses its ability to capture meaningful degradation representations, which directly leads to a significant drop in enhancement performance. 
We attribute this to the fact that, without pretraining, the DAM is treated merely as a generic feature extractor rather than a degradation-oriented encoder, thus losing its capacity to represent degradation effectively.

\begin{table}[!ht]
    \renewcommand\arraystretch{1.3}
    \centering
    \caption{Enhancement performance with and without DAM pretraining. }
    \begin{tabular*}{\hsize}{@{}@{\extracolsep{\fill}}cccccc@{}}
        \hline
        \multirow{2}{*}{Setup} & \multicolumn{2}{c}{SCARED-noise} &  & \multicolumn{2}{c}{SES} \\ \cline{2-3} \cline{5-6} 
                               & PSNR$\uparrow$         & SSIM$\uparrow$        &  & NIQE$\downarrow$       & PIQE$\downarrow$       \\ \hline
        DGGAN w/o p            & 27.21        & 0.8552      &  & 6.19       & 22.87      \\
        DGGAN                  & 33.21        & 0.9057      &  & 3.62       & 13.79      \\ \hline
    \end{tabular*}
    \label{tab:degradation_representation}
\end{table}

\subsubsection{Effectiveness of the degradation propagation}

In this part, we study the impact of different propagation intervals on model performance and inference efficiency. 
In our default setting, the propagation interval $\Delta_T$ is set to 15. 
To further examine its influence, we test the model with $\Delta_T$ values of 3, 5, 10, 20, and 30, and report the average performance and efficiency in Table \ref{tab:degradation_propagation}. 
The results show that smaller $\Delta_T$ values yield better enhancement quality but lower inference efficiency, whereas larger $\Delta_T$ values achieve the opposite trade-off. 
This observation is consistent with our intuition: shorter intervals allow more accurate degradation estimation, while longer intervals reduce computational redundancy. 
Considering that endoscopic videos are typically recorded at 30 fps, we choose $\Delta_T=15$ as a balance, which ensures satisfactory enhancement quality while maintaining real-time performance.

\begin{table}[!ht]
    \renewcommand\arraystretch{1.3}
    \centering
    \caption{The enhancement performance under different propagation intervals $\Delta_T$. }
    \begin{tabular*}{\hsize}{@{}@{\extracolsep{\fill}}cccccccc@{}}
        \hline
        \multirow{2}{*}{$\Delta_T$} & \multicolumn{2}{c}{SCARED-noise} &  & \multicolumn{2}{c}{SES} &  & \multirow{2}{*}{\begin{tabular}[c]{@{}c@{}}Running\\ time (s)\end{tabular}} \\ \cline{2-3} \cline{5-6} 
                 & PSNR$\uparrow$         & SSIM$\uparrow$        &  & NIQE$\downarrow$       & PIQE$\downarrow$          &  & \\ \hline
        3        & 32.54        & 0.8844      &  & 3.64       & 14.06         &  & 0.0973 \\
        5        & 32.18        & 0.8803      &  & 3.70       & 14.49         &  & 0.0525 \\
        10       & 31.79        & 0.8776      &  & 3.77       & 15.13         &  & 0.0408 \\
        15       & 31.03        & 0.8678      &  & 3.98       & 16.82         &  & 0.0369 \\
        20       & 30.18        & 0.8595      &  & 4.09       & 17.92         &  & 0.0350 \\
        30       & 28.32        & 0.8391      &  & 4.27       & 19.03         &  & 0.0330 \\ \hline
    \end{tabular*}
    \label{tab:degradation_propagation}
\end{table}

\subsubsection{Effectiveness of the cyclical consistency}

In this study, we analyze the contribution of each component in the cyclical consistency loss. 
As defined in \eqref{equ:cyclical_loss}, the cyclical loss consists of three terms, $L_{cl}$, $L_{ch}$, and $L_{cd}$. 
We conduct ablation experiments by removing each of these terms individually and report the results in Table \ref{tab:cyclical_consistency}. 
The results show that $L_{cl}$ and $L_{ch}$ have a major impact on the model's performance, and removing either of them leads to a substantial drop in enhancement quality. 
In contrast, the role of $L_{cd}$ is relatively minor, as it mainly penalizes inconsistency in degradation representation. 
Interestingly, removing $L_{cd}$ only slightly affects the final performance, suggesting that the primary benefits of cyclical consistency arise from the content- and quality-level constraints rather than the degradation-level constraint.

\begin{table}[!ht]
    \renewcommand\arraystretch{1.3}
    \centering
    \caption{The effectiveness of the cyclical consistency loss.}
    \begin{tabular*}{\hsize}{@{}@{\extracolsep{\fill}}cccccc@{}}
        \hline
        \multirow{2}{*}{Setup} & \multicolumn{2}{c}{SCARED-noise} &  & \multicolumn{2}{c}{SES} \\ \cline{2-3} \cline{5-6} 
                            & PSNR$\uparrow$         & SSIM$\uparrow$        &  & NIQE$\downarrow$       & PIQE$\downarrow$       \\ \hline
        w/o $L_{cl}$        & 21.53        & 0.5210      &  & 9.38       & 22.81      \\
        w/o $L_{ch}$        & 18.87        & 0.4983      &  & 11.01      & 25.75      \\
        w/o $L_{cd}$        & 32.77        & 0.8842      &  & 4.03       & 15.80      \\
        Ours                & 33.21        & 0.9057      &  & 3.62       & 13.79      \\ \hline
    \end{tabular*}
    \label{tab:cyclical_consistency}
\end{table}

\section{Conclusion}

We developed and validated a novel endoscopic video enhancement framework. 
Our approach estimates degradation representations of images and propagates them across video frames, thereby enabling high-quality real-time enhancement of endoscopic videos. 
By achieving a superior balance between performance and efficiency compared with several state-of-the-art methods, our work demonstrates the value of degradation representations in image enhancement. 
Moreover, this degradation-guided enhancement paradigm implies that, for endoscopic images in a specific minimally invasive surgery, it is sufficient to evaluate and model the potential degradations in order to train an enhancement model, which can then deliver high-quality real-time enhancement during surgery.

Nevertheless, our method has certain limitations. 
On the one hand, to ensure real-time performance, the parameter size of the single-frame enhancement module was reduced, leading to a performance gap compared with the best-performing state-of-the-art methods. 
On the other hand, our current design employs a fixed interval for degradation feature estimation, which may limit optimal performance. 
In future work, we plan to explore a cascaded inference pipeline, where image quality will be monitored in real time to dynamically decide whether degradation estimation is necessary, rather than enforcing estimation at fixed intervals.

\section*{Acknowledgments}
The National Key Research and Development Program of China supports this work under Grant 2022YFB4703000.

\bibliographystyle{unsrt}  
\bibliography{references}

\end{document}